\documentclass[conference]{IEEEtran}
\IEEEoverridecommandlockouts
\usepackage{cite}
\usepackage{amsmath,amssymb,amsfonts}
\usepackage{algorithmic}
\usepackage{graphicx}
\usepackage{textcomp}
\usepackage{float}
\usepackage{url}   
\usepackage{multirow}
\usepackage{xcolor}
\usepackage[colorlinks=true]{hyperref}
\newcommand{\redref}[1]{{\color{red}\ref{#1}}} 

\def\BibTeX{{\rm B\kern-.05em{\sc i\kern-.025em b}\kern-.08em
    T\kern-.1667em\lower.7ex\hbox{E}\kern-.125emX}}
    
\begin{document}

\title{Non-Stationary Time Series Forecasting Based on Fourier 
Analysis and Cross Attention Mechanism\\
}

\author{
\IEEEauthorblockN{Yuqi Xiong, Yang Wen\textsuperscript{*}}
\IEEEauthorblockA{\textit{College of Electronics and Information Engineering},
\textit{Shenzhen University}\\
Shenzhen, China \\
2022090048@email.szu.edu.cn, wen\_yang@szu.edu.cn}
\thanks{*Corresponding Author}
}

\maketitle

\begin{abstract}
Time series forecasting has important applications in financial analysis, weather forecasting, and traffic management. However, existing deep learning models are limited in processing non-stationary time series data because they cannot effectively capture the statistical characteristics that change over time. To address this problem, this paper proposes a new framework, AEFIN, which enhances the information sharing ability between stable and unstable components by introducing a cross-attention mechanism, and combines Fourier analysis networks with MLP to deeply explore the seasonal patterns and trend characteristics in unstable components. In addition, we design a new loss function that combines time-domain stability constraints, time-domain instability constraints, and frequency-domain stability constraints to improve the accuracy and robustness of forecasting. Experimental results show that AEFIN outperforms the most common models in terms of mean square error and mean absolute error, especially under non-stationary data conditions, and shows excellent forecasting capabilities. This paper provides an innovative solution for the modeling and forecasting of non-stationary time series data, and contributes to the research of deep learning for complex time series. Our code is publicly available at {{\hypersetup{urlcolor=blue}\url{https://github.com/YukiBear426/AEFIN}}}.  

\end{abstract}


\section{Introduction}

Time series forecasting is crucial in various practical fields, including financial analysis, weather forecasting, and traffic flow management. In recent years, with the advancement of deep learning technologies, architectures such as the Transformer have garnered increasing attention. Using the attention mechanism, the Transformer can automatically capture relationships between elements in a sequence, which has led to significant success in applications in natural language processing \cite{b1}, computer vision \cite{b2}, and speech processing \cite{b3}. Specifically, in the time series forecasting task we are addressing \cite{b4}, the Transformer has made notable strides in modeling complex time dynamics, making it highly effective for tasks involving long-term dependencies and multivariate sequences. Moreover, as powerful variants of the Transformer, models like Informer\cite{b5}, Autoformer\cite{b6}, and FEDformer\cite{b7} have demonstrated their superior capabilities in processing time series data.

\begin{figure}[b]
    \centering
    \includegraphics[width=1\linewidth]{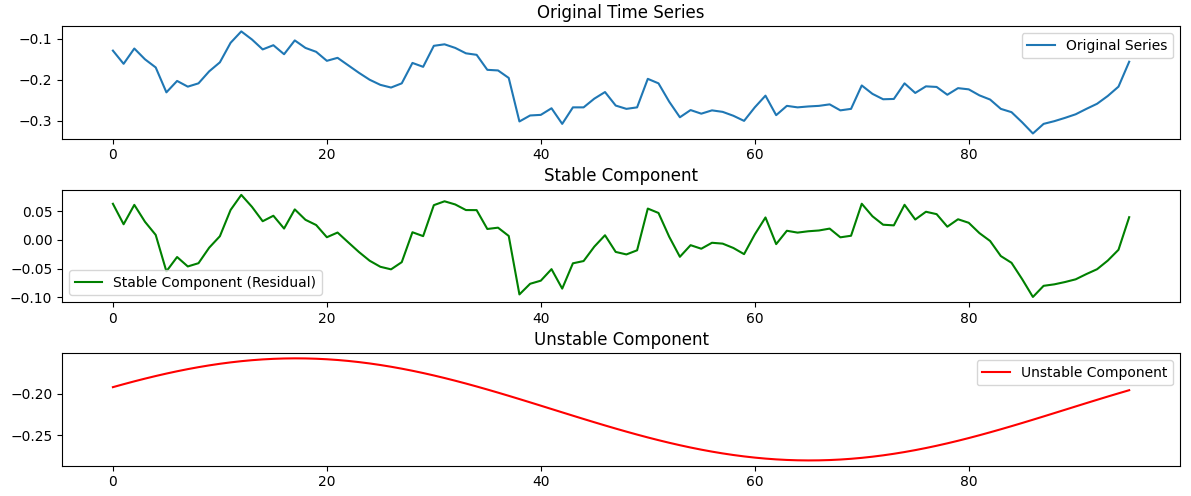}
    \caption{The original sequence, unstable components, and stable components at the first 100 time lengths in ExchangeRate dataset.}
    \label{fig:example1}
    \label{fig:enter-label1}
\end{figure}

Although these models perform well when processing stationary time series data, their performance is often severely affected by non-stationary time series data that are prevalent in the real world. The statistical properties of non-stationary time series data, such as mean and variance, change over time. Fig.\ref{fig:example1} shows the original sequence, unstable components, and stable components at the first 100 time lengths in a dataset. This instability poses a major challenge to the generalization ability of deep models and leads to a decrease in forecasting accuracy.

In existing studies, researchers have proposed a variety of methods to alleviate the impact of non-stationary characteristics on time series forecasting. One effective method is to reduce the non-stationary characteristics of data through normalization. For example, RevIN \cite{b8} can restore the non-stationary information of the time series by normalizing the input time series within the window and performing denormalization after forecasting. This method can reduce the instability of the data and improve the generalization ability of the model. However, it may lead to the problem of over-stationarization, which makes the forecasting model seriously overfitted and hinders the forecasting ability of the model. SAN \cite{b9} processes the statistical characteristics of each slice by time slicing and reduces the difficulty of the problem by decomposing the non-stationary forecasting task into two subtasks: statistical forecasting and stationary forecasting. However, the statistical forecasting module and the time series forecasting module are trained in stages, which may ignore the potential collaborative relationship between the two and limit the full exploitation of the model performance. NS-transformer \cite{b10} improves the forecasting capability of time series through series stationarization and restores the intrinsic non-stationary characteristics of the series through attention-based destationarization. However, sometimes non-stationary characteristics may be the main source of noise, and forced recovery may reduce the forecasting accuracy. FAN \cite{b11} introduces a new method called frequency adaptive normalization, which uses Fourier transform to identify the instance dominant frequency components covering most non-stationary factors to divide the series into stable and unstable components and use a simple multi-layer perceptron to predict the non-stationary components. At the same time, the current common time series forecasting model and its successful variants are used to predict the stable components. However, these methods may ignore the interdependence between components when separating and reconstructing unstable components, which limits the model’s in-depth understanding of the intrinsic structure of the time series. At the same time, using a simple multi-layer perceptron to predict unstable components will greatly weaken the use of seasonal and trend characteristics.

To address this problem, this study proposes a new model framework, Attention-Enhanced Fourier-Integrated Network (AEFIN). It introduces a cross-attention mechanism \cite{b12} to enhance the information sharing between stable and unstable components, targeting the problem of inter-component interdependence that may be ignored by existing models when processing unstable information. In addition, to address the shortcomings of existing models in capturing unstable information of time series, we propose a neural network combined with Fourier analysis to process non-stationary components. This strategy not only focuses on trend features through multi-layer perceptrons, but also deeply explores seasonal patterns through Fourier analysis networks, allowing our model to reconstruct non-stationary components more accurately. The mean square error and mean absolute error of our proposed model on multiple datasets and forecasting time windows surpass the baselines of existing major models, highlighting its robustness and reliability in capturing time dependencies, especially in the presence of non-stationary data.

The purpose of this study is to propose a new deep learning framework to improve the forecasting accuracy of non-stationary time series data and fill the gaps in existing methods in dealing with non-stationarity. Our contributions mainly include:

\begin{itemize}
    \item Use the cross-attention mechanism to enhance the information sharing between stable and unstable components in the sequence;
    \item Combine the Fourier analysis network with the multi-layer perceptron to enhance the model's ability to capture seasonality and trends;
    \item Design a new loss function combining stability constraints in both the time domain and the frequency domain;
\end{itemize}

\section{Related Work}

\subsection{Attention Mechanism in Time Series Forecasting}

\subsubsection{Temporal Domain Attention Model}

The extended model of the basic transformer model dynamically assigns attention weights through the self-attention mechanism to learn the dependencies between each element in the sequence and other elements. However, the original Transformer model is highly complex and cannot efficiently process long sequence data. To this end, researchers have proposed a variety of optimization methods. For example, Informer \cite{b5} proposed the ProbSparse mechanism based on sparse self-attention, which significantly reduced the computational complexity by selecting important attention values. This approach accelerates model inference while preserving key timing information. Autoformer \cite{b6} takes time series decomposition as its core module and introduces an autocorrelation mechanism that can adaptively capture the trend and cyclical dependencies of time series. In this way, the model can not only model dependencies within the sequence, but also better handle the nonlinear dynamics of the sequence. However, as far as we know, most of the actual data have non-stationary factors. For non-stationary time series, directly applying the above attention mechanism is often difficult to effectively capture patterns with drastic changes. Furthermore, NS-Transformer \cite{b10} proposed an attention module that combines stationarity and non-stationarity in series to specifically address the problem of over-smoothing and improve the model's adaptation to non-stationary temporal patterns. Although these models have made progress in time series analysis, they do not pay enough attention to the information exchange between the unstable and stable components in the time series. To enhance the information sharing between stable and non-stationary components in a sequence, we propose to use a cross-attention mechanism for the stable and non-stationary components in a time series. In this way, the model can more effectively capture and utilize the unstable and stable information in time series data, and improve the ability to handle non-stationary time series forecasting.

\subsubsection{Frequency Domain Attention Model}

In addition to time domain modeling, the frequency domain characteristics of time series also contain important information. FEDformer \cite{b7} uses Fourier transform and wavelet analysis techniques to propose an enhancement module based on multi-wavelet neural operator learning. This method captures the important structural features in the sequence through frequency domain mapping and significantly improves the forecasting effect by combining it with time domain feature modeling. ETSformer \cite{b13} focuses on extracting the most important frequency domain information and designs an attention mechanism to select the top-K maximum values, further reducing redundant computation and attention allocation to irrelevant frequency components. This selective attention strategy enables the model to focus more on key patterns and improves the modeling ability of complex time series. In the field of exploring non-stationary signal processing and time series analysis, a variety of methods based on Fourier transform and neural networks have been proposed to improve the performance and efficiency of the model. Based on the Fourier neural operator, AFNO \cite{b14} introduced an effective token mixing mechanism that can perform feature mixing in the Fourier domain, thereby enhancing the model's ability to represent time series data. FNet \cite{b15} replaces the traditional self-attention operation by applying the Fourier transform to the self-attention mechanism. This approach not only improves the computational efficiency of the model, but also maintains high accuracy in multiple benchmarks of natural language processing, showing its effectiveness in processing sequence data. T-WaveNet \cite{b16} constructs a tree-structured network in which each node is iteratively decomposed using a wavelet transform unit based on a reversible neural network. This structure enables the model to process and analyze the hierarchical structure of signals more efficiently. AWT-Net \cite{b17} uses wavelet transform coefficients to divide each signal point into high-frequency and low-frequency sub-band components. In this way, the model is able to capture the multi-scale characteristics of the signal and leverage the Transformer architecture to enhance the representation of the original shape features.

\subsection{Non-Stationary Time Series Forecasting}

As discussed above, the treatment of unstable ingredients in time series is also one of the key areas of time series forecasting. In order to solve the problem of instability, many methods use different modeling techniques to directly model non-stationary phenomena. DDG-DA \cite{b18} adopts a domain adaptation paradigm to predict future data distribution by establishing an effective mapping relationship between the source domain and the target domain, thereby helping the model to quickly adapt and maintain high performance when unstable factors appear. Adarnn \cite{b19} proposed an adaptive RNN model to mitigate the impact of data non-stationarity on forecasting performance using a distribution-matching mechanism. NS-Transformer \cite{b10} can effectively deal with non-stationarity in time series by introducing a dynamic stationarity modeling mechanism, that is, the statistical characteristics of data change over time. It adaptively adjusts the learning method of time dependencies and can flexibly capture trends, seasonality, and change patterns in data, thereby improving forecasting accuracy. All of the above articles assume that non-stationarity in time series can be captured by a fixed pattern or mechanism. However, in the face of highly complex and changeable time series data in reality, the adaptability of these methods may be limited. In order to model more complex nonlinear dynamic systems, global and local Koopman operators are used to learn dynamic patterns at different scales, thereby achieving multi-scale analysis and modeling of system states. At the same time, the Koopman operator combined with the Fourier transform \cite{b20} to identify the nonstationary of system models has also made important progress in the modeling and forecasting of nonstationary time series. However, these solutions are still unable to flexibly adapt to the changes in the frequency components of the input signal, which limits their applicability in dealing with complex dynamic systems with time-varying frequencies.

\subsection{Fourier Analysis Network}

Fourier analysis is a powerful tool for processing time series data, especially in capturing periodic and seasonal patterns. Traditional Fourier analysis methods identify periodic components by decomposing time series into sine and cosine functions of different frequencies \cite{b21}. However, these methods usually require time series data to be periodic and stationary, and have limited analytical capabilities for non-stationary time series. In recent years, Fourier transform-based deep learning methods, such as Fourier neural networks \cite{b22}, have combined Fourier transform with neural networks to capture the frequency information of time series data. Inspired by the Fourier analysis network, the module for predicting non-stationary factors in our AEFIN model combines multi-layer perceptrons and Fourier analysis networks, so that it can not only focus on trend characteristics, but also deeply explore seasonal patterns, thereby more accurately reconstructing non-stationary components.

\section{Methodology}

\begin{figure*}[!htbp]
    \centering
    \includegraphics[width=\linewidth]{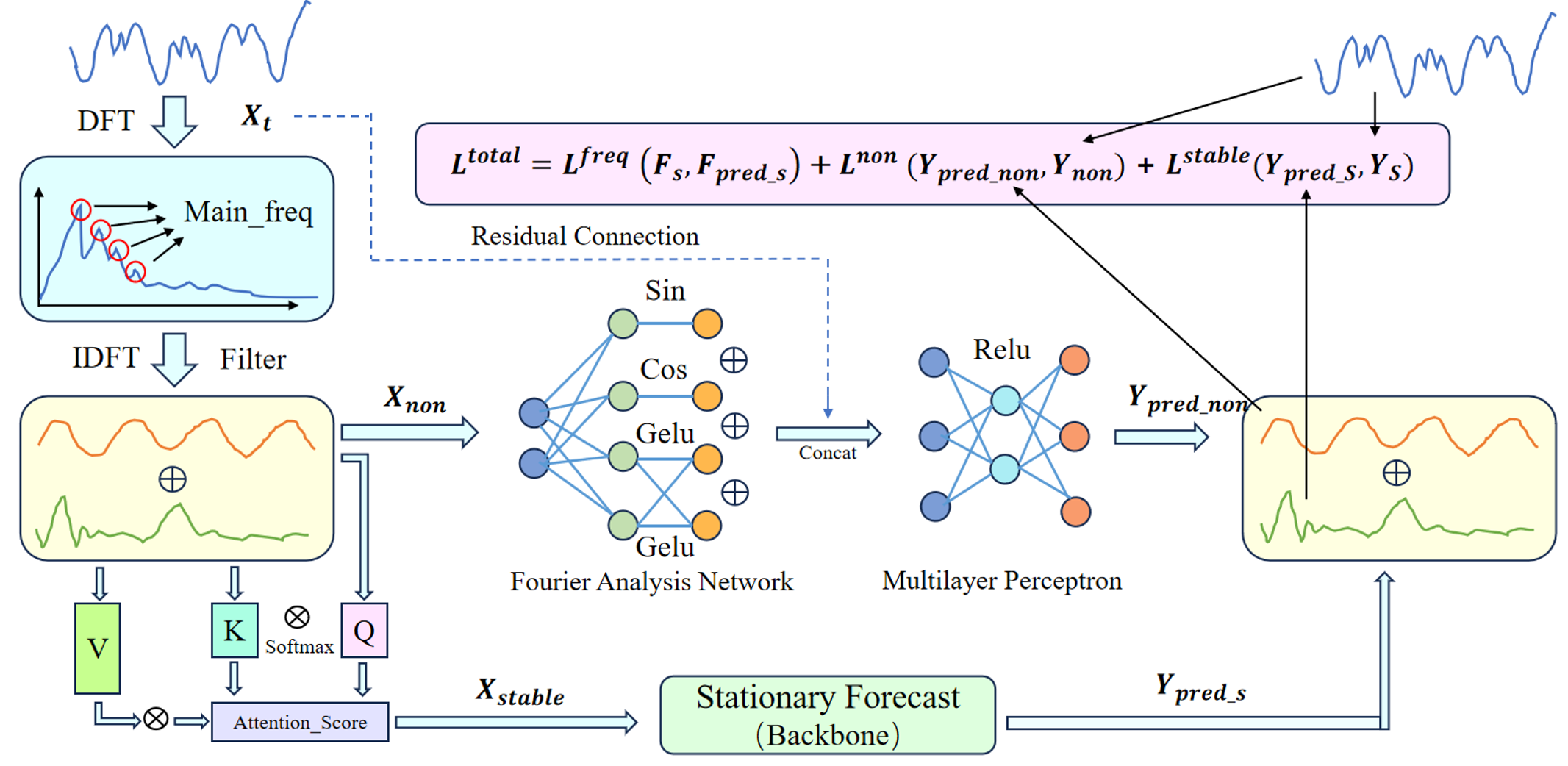}
    \caption{The overall architecture of our model AEFIN. $X_t$ is the input data, the blue arrow is the direction of data flow, the blue dashed line indicates the direction of flow of the data used for residual connection, and the black arrow indicates the source of the data in the loss function.} 
    \label{fig:example2}
    \label{fig:label_name}
\end{figure*}

Given a multivariate time series \( x \in \mathbb{R}^{N \times D} \), where \( N \) is the total time step and \( D \) is the feature dimension. We use an input sequence of length \( L \) to predict an output sequence of length \( H \). The forecasting task can be expressed as \(x_{t-L:t} \to x_{t+1:t+H},\) where \( x_{t-L:t} \in \mathbb{R}^{L \times D} \) and \( x_{t+1:t+H} \in \mathbb{R}^{H \times D} \). For a clearer representation, we denote the input and output series as \( X_t \) and \( Y_t \) respectively. Our model first performs a separation process, separating the stationary component and the non-stationary component through frequency domain decomposition, and then performs cross-attention calculation to obtain a stable component output with the fusion information of the two. The dimension remains \( X_t \in \mathbb{R}^{L \times D} \). Finally, the two components are predicted separately, and the results are merged to obtain the final output. The overall framework of the proposed method is illustrated in Fig.\ref{fig:example2}.

\subsection{Decomposition Method Based on Frequency Domain}

First, AEFIN removes the first $K$ dominant components in the frequency domain of each input instance, so that the forecasting backbone can focus on the stationary aspects of the input. We call this process frequency residual learning. At the time $t$ length, the input $X_t$ is a time series with $D$ dimensional features. However, changes in one dimension may take some time to appear in another dimension, so each feature dimension may have a different frequency pattern. Therefore, we use frequency residual learning independently in the feature channel dimension. The specific process of frequency residual learning is achieved by performing a one-dimensional discrete Fourier transform on each input $X_t$, as shown in Formula 1 and 2.

\begin{equation}
Z_t = \text{DFT}(X_t) \quad K_t = \text{Top} \, K(Z_t)
\label{eq:formula1}
\end{equation}

\begin{equation}
X_{\text{non\_stable}} = \text{IDFT}(\text{Filter}(Z_t, K_t))
\label{eq:formula2}
\end{equation}

Formula 1 and 2 shows that the discrete Fourier transform converts the Fourier components of the input to the frequency domain, recorded as $Z_t \in \mathbb{C}^{t \times D}$. Then, $\text{Top}K$ selects the frequencies with the largest amplitudes in the first $K$ and records their positions in the sequence. Then, the previously recorded position information is used from $Z_t$, and $K_t$ is filtered out using a filter to obtain the frequency value of the unstable component. Next, the frequency domain of the unstable component is subjected to the inverse discrete Fourier transform to obtain the unstable component $X_{\text{non\_stable}}$ separated in the time domain. At this time, the dimension returns to $X_{\text{non\_stable}} \in \mathbb{R}^{L \times D}$. Then, the stationary component can be obtained by removing $X_{\text{non\_stable}}$ from the original sequence $X_t$, as shown in Formula 3.

\begin{equation}
X_{\text{stable}} = X_t - X_{\text{non\_stable}}
\label{eq:formula3}
\end{equation}

In the entire frequency residual learning process, both the discrete Fourier transform and the inverse discrete Fourier transform can be calculated using the fast Fourier transform\cite{b23}, with a computational complexity of $O(L \log L)$. The complexity of the filtering operation of selecting $K$ dominant components, $\text{Top}K$ and $\text{Filter}$ is $O(L + K)$.

\subsection{Information Fusion Based on Cross Attention}

After the previous frequency domain decomposition, we get the stable component $X_{\text{stable}}$ and the unstable component $X_{\text{non\_stable}}$ in the sequence, with dimensions both $X \in \mathbb{R}^{L \times D}$. We now use the unstable component as the Query ($Q$), and the stable component as the Key ($K$) and Value ($V$).

The core of cross attention is to calculate the attention weight by $Q$ and $K$. For each query vector $q_{\text{non\_stable}} \in \mathbb{R}^D$, we calculate its similarity with all key vectors $k_{\text{stable}} \in \mathbb{R}^D$. The specific calculation method is given by Formula 4.

\begin{equation}
{M}_{i,j} = \frac{q_{\text{non\_stable}}^T k_{\text{stable}}}{\sqrt{D}}
\label{eq:formula32}
\end{equation}

where ${M}_{i,j}$ is the attention score matrix, $q_{\text{non\_stable}}$ is the query vector from $X_{\text{non\_stable}}$, $k_{\text{stable}}$ is the key vector from $X_{\text{stable}}$, and $D$ is the dimension of the vector.

The Softmax operation is then applied to the attention score to obtain the attention weight between query and key, as shown in Formula 5, where $\alpha_{i,j}$ is the attention weight between query $i$ and key $j$.

\begin{equation}
\alpha_{i,j} = \frac{\exp({M}_{i,j})}{\sum_{k=1}^{L} \exp({M}_{i,k})}
\label{eq:formula4}
\end{equation}

The output representation is obtained by weighted summing the attention weight with the value $v_{\text{stable}} \in \mathbb{R}^D$ corresponding to $X_{\text{stable}}$, as shown in Formula 6.

\begin{equation}
o_i = \sum_{j=1}^L \alpha_{i,j} v_{\text{stable}}
\label{eq:formula5}
\end{equation}

Among them, the output $o_i$ is the weighted sum of all stable components $v_{\text{stable}}$. All outputs are aggregated into a matrix $O \in \mathbb{R}^{L \times D}$, which is the stable output $X_{\text{new\_stable}} \in \mathbb{R}^{L \times D}$ that integrates the information of unstable components. 

In order to enable the stable and unstable components to better extract information from the fragments and input them into the subsequent neural network, we slice $X_{\text{stable}}$ and $X_{\text{non\_stable}}$ to obtain $batch$. At this time, their shapes become $\mathbb{R}^{B \times C \times L_{\text{in}}}$, where $B$ is the batch size, $C$ is the number of channels, and $L_{\text{in}}$ is the input length.

\subsection{Fourier Analysis Network}

\subsubsection{Theoretical basis}
We know that a periodic function can be expressed using the Fourier series expansion form, as shown in Formula 7.

\begin{equation}
f_{(x)} = a_0 + \sum_{n=1}^{\infty} \left(a_n \cos\left(\frac{2 \pi n x}{T}\right) + b_n \sin\left(\frac{2 \pi n x}{T}\right)\right)
\label{eq:formula6}
\end{equation}

Among them, $a_0$ is a constant term, $a_n$ and $b_n$ are Fourier coefficients, representing the amplitudes of cosine and sine respectively, and $T$ is the period. In order to transform the expression of this Fourier series into a neural network model, the Fourier series is expanded into a matrix form. By introducing matrix multiplication, the Fourier series can be expressed as Formula 8.

\begin{equation}
fS_{(x)} = B + \sum_{n=1}^{N} \left(W_{c}[n] \cos(W_{in} X) + W_{s}[n] \sin(W_{in} X)\right)
\label{eq:formula7}
\end{equation}

Here, $B$ is a constant term corresponding to $a_0$ in the Fourier series, $W_c$ and $W_s$ represent the learnable weights of the cosine and sine terms, respectively. $W_{in}$ is the input weight matrix, which is multiplied by the input $X$ to adjust the frequency. This transformation allows the Fourier series to be calculated through matrix multiplication. Further simplification gives Formula 9.

\begin{equation}
fS_{(x)} = B + W_{c} \cos(W_{in} X) + W_{s} \sin(W_{in} X)
\label{eq:formula8}
\end{equation}

Among them, $W_c$ and $W_s$ are learnable matrices that control the cosine and sine parts of each layer of Fourier series. In order to improve the expression ability, we can splice the cosine and sine parts together into a unified matrix expression, such as Formula 10.

\begin{equation}
fS_{(x)} = B + W_{out} \left[ \cos(W_{in} X) \parallel \sin(W_{in} X) \right]
\label{eq:formula9}
\end{equation}

Where $W_{out}$ is a new output weight matrix. This form allows the Fourier transform result to be obtained through simple matrix operations, and then it can be stacked in multiple layers like the multi-layer perceptron we are familiar with. However, directly stacking such a network will cause the model to focus on the frequency part, that is, $W_{in}$, during the learning process and ignore the learning of the Fourier coefficients $a_n$ and $b_n$. By stacking multiple Fourier transform layers, the network learns more about the frequency adjustment, while the Fourier coefficients are not effectively learned. To solve this problem, we need to design a network so that the learning ability of the Fourier coefficients is positively correlated with the network depth and the intermediate layers can also effectively model the periodic features. Therefore, we first process the input through a Fourier transform, and then enhance the nonlinear expression ability through the activation function. In this way, the outputs of the cosine, sine, and activation functions are spliced together, which increases the network's expression ability and can better handle periodic data. The specific calculation is shown in Formula 11.

\begin{equation}
\varphi(x) = \left[\cos\left(W_p X\right) \parallel \sin\left(W_p X\right) \parallel \sigma\left(B_p + W_p X\right)\right]
\label{eq:formula10}
\end{equation}

Where $W_p$ is the input weight of the Fourier transform, $\sigma$ is the nonlinear activation function, and $B_p$ is the bias term.

\subsubsection{Specific implementation}
Since different feature channels may have different periodic transformations, we separate the unstable components $X_{\text{non\_stable}} \in \mathbb{R}^{B \times C \times L_{\text{in}}}$ in the channel dimension to obtain a sequence of $C$ dimensions of $\mathbb{R}^{B \times 1 \times L_{\text{in}}}$, and then reduce its dimension to obtain a sequence of $C$ dimensions of $\mathbb{R}^{B \times L_{\text{in}}}$. Then, they pass through two parallel fully connected layers to obtain outputs $O_1$ and $O_2$, as shown in Formula 12.

\begin{equation}
O_1 = W_1 X + b_1 \quad O_2 = \text{GELU}(W_2 X + b_2)
\label{eq:formula11}
\end{equation}

Where $X \in \mathbb{R}^{B \times L_{\text{in}}}$ is the input, $W_1 \in \mathbb{R}^{L_{\text{pred}//4} \times L_{\text{in}}}$ is one of the weight matrices, $W_2 \in \mathbb{R}^{L_{\text{pred}//2} \times L_{\text{in}}}$ is another weight matrix, $b_1 \in \mathbb{R}^{L_{\text{pred}//4}}$ is one of the bias terms, $b_2 \in \mathbb{R}^{L_{\text{pred}//2}}$ is another bias term, and $\text{GELU}$ is the activation function, as shown in Formula 13 and 14.

\begin{equation}
\text{GELU}(x) = 0.5 x \left( 1 + \tanh{\left( \frac{x}{\sqrt{2}} + 0.04475 x^3 \right)} \right)
\label{eq:formula12_1}
\end{equation}

\begin{equation}
\text{tanh}(x) = \frac{e^x - e^{-x}}{e^x + e^{-x}}
\label{eq:formula12_2}
\end{equation}

$O_1 \in \mathbb{R}^{B \times L_{\text{pred}//4}}$ and $O_2 \in \mathbb{R}^{B \times L_{\text{pred}//2}}$ are the outputs of the two layers respectively. Then the cosine and sine values of the output of $O_1$ without the activation function are calculated, as shown in Formula 15.

\begin{equation}
O_{\sin} = \sin(O_1) \quad O_{\cos} = \cos(O_1)
\label{eq:formula13}
\end{equation}

The dimensions of the outputs $O_{\sin}$ and $O_{\cos}$ are the same as $O_1$, which is $\mathbb{R}^{B \times L_{\text{pred}//4}}$. Next, the three outputs $O_{\sin}$, $O_{\cos}$, and $O_2$ are concatenated in the time dimension to obtain the predicted output $Y_{\text{pred}} \in \mathbb{R}^{B \times L_{\text{pred}}}$ of each dimension, as shown in Formula 16.

\begin{equation}
Y_{\text{pred}} = \text{Concat}(O_{\cos}, O_{\sin}, O_2)
\label{eq:formula14}
\end{equation}

Finally, the predicted output $Y_{\text{pred}}$ of each dimension is upgraded to obtain $\mathbb{R}^{B \times 1 \times L_{\text{pred}}}$, and then merged in the channel dimension to obtain the final output $Y_{\text{pred}} \in \mathbb{R}^{B \times C \times L_{\text{pred}}}$. For convenience in the discussion in the following chapters, we will use the symbol FAN to replace the above Fourier analysis network forecasting part.

\subsection{Forecasting Process}
From the previous analysis, we get the unstable component \( X_{\text{non\_stable}} \) in the time series and the stable component \( X_{\text{new\_stable}} \), which is added and fused with unstable information through cross-attention. Their dimensions are \( R^{B \times C \times L_{\text{in}}} \). Inputting the stable component into the forecasting backbone model Backbone can get the forecasting output of the stable part, as shown in Formula 17.

\begin{equation}
Y_{\text{pred\_stable}} = \text{Backbone}(X_{\text{new\_stable}})
\label{eq:formula15}
\end{equation}

Then we predict the unstable part. Since the unstable component contains a lot of seasonal and trend information, we first let the stable component pass through the Fourier analysis network to extract the seasonal information in the sequence, as shown in Formula 18.

\begin{equation}
Z_1 = \text{FAN}(X_{\text{non\_stable}})
\label{eq:formula16}
\end{equation}

After obtaining the output \( Z_1 \in R^{B \times C \times L_{\text{pred}}} \), it is concatenated with the original sequence \( X \in R^{B \times C \times L_{\text{in}}} \) in the last dimension, as shown in Formula 19.

\begin{equation}
Z_2 = \text{Concat}(Z_1, X)
\label{eq:formula17}
\end{equation}

The concatenated output \( Z_2 \in R^{B \times C \times (L_{\text{pred}} + L_{\text{in}})} \) is then passed through a simple multi-layer perceptron to extract trend information. The first layer of the perceptron is as shown in Formula 20.

\begin{equation}
Z_3 = \text{ReLU}(Z_2 W_1 + b_1)
\label{eq:formula18}
\end{equation}

Where \( W_1 \in R^{(L_{\text{pred}} + L_{\text{in}}) \times 3(L_{\text{pred}} + L_{\text{in}})} \) is the weight matrix of the first layer, and \( b_1 \in R^{3(L_{\text{pred}} + L_{\text{in}})} \) is the bias term of the first layer. Then, after passing through the second layer of the perceptron, as shown in Formula 21.

\begin{equation}
Y_{\text{pred\_nonstable}} = Z_3 W_2 + b_2
\label{eq:formula19}
\end{equation}

Where \( W_2 \in R^{3(L_{\text{pred}} + L_{\text{in}}) \times L_{\text{pred}}} \) is the weight matrix of the second layer, and \( b_2 \in R^{L_{\text{pred}}} \) is the bias term of the second layer. At this time, the forecasting output of the unstable component \( Y_{\text{pred\_nonstable}} \) and the forecasting output of the stable component \( Y_{\text{pred\_stable}} \) are both \( R^{B \times C \times L_{\text{pred}}} \), so that the final forecasting output is obtained by adding the two in the time dimension, as shown in Formula 22.

\begin{equation}
Y_{\text{pred}} = Y_{\text{pred\_stable}} + Y_{\text{pred\_nonstable}}
\label{eq:formula20}
\end{equation}

\subsection{Loss Function}

\subsubsection{Stable Components in the Time Domain}

Since stable components usually represent relatively stable or low-frequency parts in time series data, they change less and are less susceptible to fluctuations. To ensure that the model accurately reconstructs the stable components in the time domain, the stable component loss in the time domain is measured by the mean square error (MSE) to measure the difference between the predicted and true stable components, as shown in Formula 23.

\begin{equation}
\mathcal{L}_{\text{stable}} = \frac{1}{N} \sum_{i=1}^{N} \left( Y_{\text{pred\_stable},i} - Y_{\text{stable},i} \right)^2
\label{eq:formula21}
\end{equation}

MSE is used as the square error of the target value to amplify the larger error, which helps to improve the accuracy of the model for stable components in the time domain, thereby ensuring that it fits the actual data as closely as possible in the time domain. In this way, the model's forecasting of stable components can more accurately conform to the real stable trend.

\subsubsection{Time-Domain Unstable Components}

Since unstable components usually represent high-frequency or highly volatile parts, they have strong uncertainty or randomness. Such components may contain noise or short-term anomalies, so we use the MAE to measure the difference between the predicted and true unstable components to capture the volatility of the component, as shown in Formula 24.

\begin{equation}
\mathcal{L}_{\text{non\_stable}} = \frac{1}{N} \sum_{i=1}^{N} \left| Y_{\text{pred\_nonstable},i} - Y_{\text{non\_stable},i} \right|
\label{eq:formula22}
\end{equation}

MAE captures the overall trend of unstable components more robustly through linear error measurement, reducing excessive penalties for outliers. Flexible treatment of unstable components helps the model to more reasonably balance the error and flexibility of volatility forecasting, thereby avoiding overfitting to short-term fluctuations in forecasting.

\subsubsection{Stability Constraint Loss in the Frequency Domain}

Since in the frequency domain, the stable component should show lower frequency components, and the model may not be able to accurately capture these characteristics due to over-optimization. The lower frequency stable components in the frequency domain usually contribute more to the overall trend. Therefore, we use the mean square error of the spectrum difference to ensure that the spectral characteristics of the predicted stable component are consistent with the true stable component, as shown in Formula 25.

\begin{equation}
\mathcal{L}_{\text{freq}} = \frac{1}{N} \sum_{i=1}^{N} \left( \left| F \left( Y_{\text{stable},i} \right) \right| - \left| F \left( Y_{\text{pred\_stable},i} \right) \right| \right)^2
\label{eq:formula23}
\end{equation}

By constraining the spectrum of stable components in the frequency domain, we can avoid the accumulation of deviations or frequency leakage in the time domain, thereby improving the structural consistency of the model in frequency. This design can help the model maintain the core spectrum characteristics of the original data in the frequency domain, avoid the predicted stable components from deviating from the overall trend, and thus enhance the model's ability to learn the long-term stable trend of the data.

\setcounter{table}{1}
\begin{table*}
\caption{Performance comparison of different models across datasets and forecasting time windows. Bold indicates the best results.} 
\label{tab:example1}
\resizebox{\textwidth}{!}{%
\begin{tabular}{cccccc|cccc|cccc}
\hline
\multicolumn{2}{c}{Method} & \multicolumn{2}{c}{DLinear} & \multicolumn{2}{c}{+AEFIN} & \multicolumn{2}{c}{Informer} & \multicolumn{2}{c}{+AEFIN} & \multicolumn{2}{c}{SCINet} & \multicolumn{2}{c}{+AEFIN} \\ 
\multicolumn{2}{c}{Metrics} & MSE & MAE & MSE & MAE & MSE & MAE & MSE & MAE & MSE & MAE & MSE & MAE \\ \hline
\multirow{4}{*}{Exchange} 
& 96  & \textbf{0.052} & \textbf{0.164} & 0.057 & 0.173 & 0.412 & 0.532 & \textbf{0.061} & \textbf{0.180} & 0.085 & 0.218 & \textbf{0.061} & \textbf{0.180} \\ 
& 168 & \textbf{0.090} & \textbf{0.219} & 0.095 & 0.229 & 0.491 & 0.582 & \textbf{0.099} & \textbf{0.237} & 0.126 & 0.266 & \textbf{0.097} & \textbf{0.232} \\ 
& 336 & \textbf{0.155} & \textbf{0.288} & 0.173 & 0.311 & 0.847 & 0.721 & \textbf{0.186} & \textbf{0.324} & 0.203 & 0.337 & \textbf{0.189} & \textbf{0.325} \\ 
& 720 & 0.352 & 0.453 & \textbf{0.296} & \textbf{0.413} & 1.210 & 0.889 & \textbf{0.356} & \textbf{0.467} & 0.430 & 0.502 & \textbf{0.342} & \textbf{0.444} \\ \hline

\multirow{4}{*}{Traffic} 
& 96  & 0.504 & 0.387 & \textbf{0.400} & \textbf{0.320} & 0.428 & 0.350 & \textbf{0.404} & \textbf{0.330} & 0.471 & 0.399 & \textbf{0.407} & \textbf{0.334} \\ 
& 168 & 0.804 & 0.588 & \textbf{0.414} & \textbf{0.334} & 0.457 & 0.366 & \textbf{0.430} & \textbf{0.347} & 0.443 & 0.377 & \textbf{0.429} & \textbf{0.356} \\ 
& 336 & 0.504 & 0.380 & \textbf{0.439} & \textbf{0.350} & 0.555 & 0.414 & \textbf{0.444} & \textbf{0.352} & \textbf{0.459} & \textbf{0.384} & 0.481 & 0.403 \\ 
& 720 & 0.532 & 0.407 & \textbf{0.478} & \textbf{0.383} & 1.002 & 0.656 & \textbf{0.736} & \textbf{0.417} & \textbf{0.490} & \textbf{0.401} & 0.558 & 0.467 \\ \hline

\multirow{4}{*}{ETTh1} 
& 96  & \textbf{0.368} & \textbf{0.424} & 0.373 & 0.437 & 0.646 & 0.598 & \textbf{0.379} & \textbf{0.443} & 0.409 & 0.461 & \textbf{0.371} & \textbf{0.436} \\ 
& 168 & \textbf{0.398} & \textbf{0.449} & 0.405 & 0.462 & 0.863 & 0.694 & \textbf{0.447} & \textbf{0.491} & 0.489 & 0.518 & \textbf{0.430} & \textbf{0.481} \\ 
& 336 & \textbf{0.448} & \textbf{0.458} & 0.451 & 0.494 & 0.950 & 0.738 & \textbf{0.541} & \textbf{0.541} & 0.582 & 0.574 & \textbf{0.500} & \textbf{0.521} \\ 
& 720 & \textbf{0.558} & \textbf{0.561} & 0.583 & 0.581 & 1.106 & 0.823 & \textbf{0.716} & \textbf{0.644} & 0.707 & 0.645 & \textbf{0.637} & \textbf{0.612} \\ \hline

\multirow{4}{*}{ETTh2} 
& 96  & \textbf{0.110} & \textbf{0.237} & 0.121 & 0.249 & 0.160 & 0.298 & \textbf{0.134} & \textbf{0.260} & 0.128 & 0.264 & \textbf{0.116} & \textbf{0.243} \\ 
& 168 & \textbf{0.127} & \textbf{0.254} & 0.146 & 0.274 & 0.191 & 0.331 & \textbf{0.152} & \textbf{0.277} & 0.156 & 0.292 & \textbf{0.141} & \textbf{0.268} \\ 
& 336 & \textbf{0.138} & \textbf{0.271} & 0.149 & 0.277 & 0.208 & 0.347 & \textbf{0.200} & \textbf{0.314} & \textbf{0.167} & 0.305 & 0.170 & \textbf{0.302} \\ 
& 720 & \textbf{0.179} & 0.316 & 0.180 & \textbf{0.311} & \textbf{0.291} & \textbf{0.413} & 0.364 & 0.440 & \textbf{0.201} & 0.339 & 0.206 & \textbf{0.337} \\ \hline
\end{tabular}%
}
\vspace{0.15cm} 
\end{table*}

\subsubsection{Total Loss}

By weighted combination of the three loss functions, we ensure that the model accurately reconstructs the time domain representation of the stable components while maintaining its robustness to the unstable components and avoiding structural deviations in the frequency domain, as shown in Formula 26.

\begin{equation}
\mathcal{L}_{\text{main}} = \alpha \mathcal{L}_{\text{stable}} + \beta \mathcal{L}_{\text{non\_stable}} + \gamma \mathcal{L}_{\text{freq}}
\label{eq:formula24}
\end{equation}

Among them, \( \alpha \), \( \beta \), and \( \gamma \) are weight coefficients. After continuous attempts in the experiment, the final values were determined to be 0.5, 0.2 and 0.3 respectively.

\section{Experiment}

\subsection{Dataset Description}

The following are five datasets commonly used in multivariate time series forecasting. ETT \cite{b5} records the oil temperature and load of power transformers from July 2016 to July 2018. The dataset includes four subsets, namely ETTh1, ETTh1 sampled every hour, and ETTm1, ETTm2 sampled every 15 minutes. Electricity contains 321 samples of electricity consumption every 15 minutes from July 2016 to July 2019. ExchangeRate contains daily exchange rates for 8 countries from 1990 to 2016. Traffic \cite{b25} includes hourly traffic loads on San Francisco highways recorded by 862 sensors from 2015 to 2016. Weather consists of 21 weather indicators, including temperature and humidity collected every 10 minutes in 2021. z-score normalization \cite{b26} is applied to the datasets used to scale different variables to the same scale. The split ratio of the training set, validation set, and test set is set to 7:2:1. We use 90\% of the maximum frequency amplitude in the training set as the main change frequency to determine the value of the hyperparameter K. Its value and related properties of the data set are shown in Table~\redref{tab:example2} below.

\setcounter{table}{0}
\begin{table}[H]
\caption{Dataset attributes and k value selection.}
\label{tab:example2}
\resizebox{8.9cm}{!}{
\begin{tabular}{ccccccc}
\hline
Dataset   & Etth1/2 & Ettm1/2 & \multicolumn{1}{l}{Electricity} & ExchangeRate & Traffic                   & Weather \\ \hline
Variables & 7       & 7       & 321                             & 8            & 862                       & 21      \\
Time step & 17420   & 69680   & 26340                           & 7588         & 17544                     & 52696   \\
K         & 4/3     & 11/5    & 3                               & 2            & 30 & 2       \\ \hline
\end{tabular}
}
\vspace{0.15cm}  
\end{table}
\vspace{-20pt}  

\subsection{Experimental Setup}

We set the forecasting length to the four most common ones {96, 168, 336, 720}, which covers both short-term and long-term forecasting. The input window is fixed to length 96. We use mean squared error and mean absolute error to evaluate the performance of the baseline. To predict the stable component in the time series, we use three of the most common time series forecasting models: DLinear based on multi-layer perceptron, Informer based on Transformer, and SCINet based on convolutional neural network \cite{b27}. For the forecasting of the unstable component in the time series, the sizes of the two hidden layers in the Fourier analysis module are one-quarter and one-half of the output respectively, and the hidden layer size of the multi-layer perceptron module is three times the input. Our experiments are run 5 times on an NVIDIA RTX 4090 GPU (24GB) with seeds \{1, 2, 3, 4, 5\} and use ADAM \cite{b28} as the optimizer.

\subsection{Experimental Results}

Table~\redref{tab:example1} shows the MAE and MSE errors of the baseline model and the AEFIN model proposed by us on the ExchangeRate, Traffic, ETTh1 and ETTh2 datasets when the forecasting step size is 96, 168, 336 and 720. Experimental results show that AEFIN combined with the Informer model can significantly outperform the baseline model in most data sets and forecasting steps, and performs particularly well on the ExchangeRate data set, with MSE improved by about 66\% to 85\% and MAE improved by about 43\% to 66\%. AEFIN combined with SCINet model can also significantly outperform the baseline model on most datasets and forecasting steps. Compared with the Dlinear baseline model, the forecasting effect of the AEFIN model is slightly inferior. Only on the Traffic dataset does our model improve the forecasting accuracy. This may be related to the linear structural characteristics of the Dlinear model, which is not suitable for predicting the stable and unstable components separately after frequency decomposition. In summary, the AEFIN model effectively extracts seasonal and trend features from unstable components and fuses the information in stable components with unstable components through the cross-attention mechanism, thus effectively dealing with time with diverse and unstable factors. When used on a large number of sequence datasets, it demonstrates remarkable robustness and contributes to the further development of time series forecasting tasks.

\subsection{Ablation Studies}

In order to evaluate the contribution of each component in our proposed model, we conducted an ablation experiment. Specifically, we tested the core modules in the model, including the cross-attention mechanism, Fourier analysis network, and loss function. By comparing the experimental results under different configurations, we explored the impact of each module on the final forecasting accuracy. The following Table~\ref{tab:example3} to Table~\ref{tab:example6} show the evaluation indicators of each module when the forecasting length is \{96, 168, 336, 720\} on the ExchangeRate dataset. The bold indicates the best effect.

\setcounter{table}{2}
\begin{table}[H]
\caption{Evaluation index of each module under 96 length.}
\label{tab:example3}
\centering
\resizebox{8.9cm}{!}{
\begin{tabular}{c c c c c c c}
\hline
Method                                                     & \multicolumn{2}{c}{DLinear}                           & \multicolumn{2}{l}{Informer}                          & \multicolumn{2}{c}{SCINet}                                        \\ \hline
metrics                                                    & MSE                       & MAE                       & MSE                       & MAE                       & MSE                                   & MAE                       \\ \hline
Baseline                                                   & \textbf{0.052}            & \textbf{0.164}            & 0.412                     & 0.532                     & 0.085                                 & 0.218                     \\ 
\begin{tabular}[c]{@{}c@{}}Cross\\ Attention\end{tabular}  & 0.054                     & 0.170                     & 0.072                     & 0.193                     & {\color[HTML]{000000} \textbf{0.055}} & \textbf{0.170}            \\ 
\begin{tabular}[c]{@{}c@{}}Fourier\\ Analysis\end{tabular} & 0.056                     & 0.169                     & 0.089                     & 0.214                     & 0.057                                 & 0.171                     \\ 
\begin{tabular}[c]{@{}c@{}}Loss\\ Function\end{tabular}    & 0.054                     & 0.169                     & \textbf{0.057}            & \textbf{0.174}            & 0.058                                 & 0.177                     \\ 
\hline
Final                                                      & 0.057                     & 0.173                     & 0.061                     & 0.180                     & 0.061                                 & 0.180                     \\ \hline
\end{tabular}
}
\vspace{0.15cm} 
\end{table}
\vspace{-20pt}

\begin{table}[H]
\caption{Evaluation index of each module under 168 length.}
\label{tab:example4}
\centering
\resizebox{8.9cm}{!}{%
\begin{tabular}{c c c c c c c}
\hline
Method                                                     & \multicolumn{2}{c}{DLinear}                           & \multicolumn{2}{l}{Informer}                                            & \multicolumn{2}{c}{SCINet}                                        \\ \hline
metrics                                                    & MSE                       & MAE                       & MSE                                & MAE                                & MSE                                   & MAE                       \\ \hline
Baseline                                                   & 0.090                     & 0.219                     & 0.419                              & 0.582                              & 0.126                                 & 0.266                     \\ 
\begin{tabular}[c]{@{}c@{}}Cross\\ Attention\end{tabular}  & 0.088                     & 0.216                     & 0.099                              & 0.240                              & {\color[HTML]{000000} \textbf{0.089}} & \textbf{0.217}            \\ 
\begin{tabular}[c]{@{}c@{}}Fourier\\ Analysis\end{tabular} & 0.093                     & 0.228                     & 0.113                              & 0.247                              & 0.090                                 & 0.224                     \\ 
\begin{tabular}[c]{@{}c@{}}Loss\\ Function\end{tabular}    & \textbf{0.085}            & \textbf{0.212}            & 0.104                              & 0.240                              & 0.095                                 & 0.225                     \\ 
\hline
Final                                                      & 0.095                     & 0.229                     & \textbf{0.099}                    & \textbf{0.237}                    & 0.097                                 & 0.232                     \\ \hline
\end{tabular}
}%
\vspace{0.15cm} 
\end{table}
\vspace{-20pt}

\begin{table}[H]
\caption{Evaluation index of each module under 336 length.}
\label{tab:example5}
\centering
\resizebox{8.9cm}{!}{%
\begin{tabular}{c c c c c c c}
\hline
Method                                                     & \multicolumn{2}{c}{DLinear}                                             & \multicolumn{2}{c}{Informer}                          & \multicolumn{2}{c}{SCINet}                               \\ \hline
metrics                                                    & MSE                                & MAE                                & MSE                       & MAE                       & MSE                          & MAE                       \\ \hline
Baseline                                                   & 0.155                              & 0.288                              & 0.847                     & 0.721                     & 0.203                        & 0.337                     \\ 
\begin{tabular}[c]{@{}c@{}}Cross\\ Attention\end{tabular}  & \textbf{0.161}            & \textbf{0.295}            & 0.257                              & 0.374                              & {\color[HTML]{000000} \textbf{0.170}} & \textbf{0.303}            \\ 
\begin{tabular}[c]{@{}c@{}}Fourier\\ Analysis\end{tabular} & 0.165                     & 0.300                     & 0.340                              & 0.425                              & 0.173                                 & 0.308                     \\ 
\begin{tabular}[c]{@{}c@{}}Loss\\ Function\end{tabular}    & 0.164                     & 0.299                     & 0.201                              & 0.388                              & 0.178                                 & 0.318                     \\ 
\hline
Final                                                      & \multicolumn{1}{l}{0.173} & \multicolumn{1}{l}{0.311} & \multicolumn{1}{l}{\textbf{0.186}} & \multicolumn{1}{l}{\textbf{0.324}} & \multicolumn{1}{l}{0.189}             & \multicolumn{1}{l}{0.325} \\ \hline
\end{tabular}
}%
\vspace{0.15cm} 
\end{table}
\vspace{-20pt}

\begin{table}[H]
\caption{Evaluation index of each module under 720 length.}
\label{tab:example6}
\centering
\resizebox{8.9cm}{!}{%
\begin{tabular}{c c c c c c c}
\hline
Method                                                     & \multicolumn{2}{c}{DLinear}                                             & \multicolumn{2}{c}{Informer}                          & \multicolumn{2}{c}{SCINet}                               \\ \hline
metrics                                                    & MSE                                & MAE                                & MSE                       & MAE                       & MSE                          & MAE                       \\ \hline
Baseline                                                   & 0.352                              & 0.453                              & 1.210                     & 0.889                     & 0.430                        & 0.502                     \\ 
\begin{tabular}[c]{@{}c@{}}Cross\\ Attention\end{tabular}  & 0.323                              & 0.424                              & 0.290                     & 0.413                     & {\color[HTML]{000000} 0.324} & 0.425                     \\ 
\begin{tabular}[c]{@{}c@{}}Fourier\\ Analysis\end{tabular} & 0.351                              & 0.449                              & \textbf{0.245}            & \textbf{0.370}            & \textbf{0.308}               & \textbf{0.414}            \\ 
\begin{tabular}[c]{@{}c@{}}Loss\\ Function\end{tabular}    & 0.297                              & 0.413                              & 0.403                     & 0.493                     & 0.333                        & 0.436                     \\ 
\hline
Final                                                      & \multicolumn{1}{l}{\textbf{0.296}} & \multicolumn{1}{l}{\textbf{0.413}} & \multicolumn{1}{l}{0.356} & \multicolumn{1}{l}{0.467} & \multicolumn{1}{l}{0.342}    & \multicolumn{1}{l}{0.445} \\ \hline
\end{tabular}
}%
\vspace{0.15cm} 
\end{table}

\subsection{Comparative Experiments}
In this section, we compare AEFIN with four state-of-the-art non-stationary time series forecasting methods: FAN \cite{b11}, SAN \cite{b9}, Dish-TS \cite{b29}, and RevIN \cite{b8}. The current experimental settings are the same as before, using the ExchangeRate dataset and the Informer model to predict the stable part. Table~\ref{tab:example7} shows the results of the comparison, which shows that our AEFIN model achieves good results, outperforming most baseline models or achieving the second-best performance. In comparison, the Dish-TS model has the worst performance, and the SAN model performs relatively well in predicting short-term time. In terms of MAE error, our model has decreased by 7.1\%; in terms of MSE error, our model has decreased by 10.9\%. In terms of long-term time forecasting, our model achieved remarkable results, improving the MAE error by 7.2\% and the MSE error by 20.5\% over the second-best model. For other time length forecasting, our model improves by 0.4\% to 2.7\% over the second-best model.

\begin{table}[H]
\caption{Comparison results, bold indicates the best result, underline indicates the second-best result.} 
\label{tab:example7} 
\centering  
\resizebox{8.9cm}{!}{  
\begin{tabular}{ccccccccc}
\hline
Length  & \multicolumn{2}{c}{96}          & \multicolumn{2}{c}{168}         & \multicolumn{2}{c}{336}         & \multicolumn{2}{c}{720}         \\ \hline
metrics & MAE            & MSE            & MAE            & MSE            & MAE            & MSE            & MAE            & MSE            \\ \hline
Revin   & 0.223          & 0.096          & 0.295          & 0.157          & 0.375          & 0.252          & \underline{0.503}    & \underline{0.448}    \\
Dish-TS & 0.278          & 0.183          & 0.364          & 0.279          & 0.566          & 0.603          & 0.730          & 0.822          \\
SAN     & \textbf{0.168} & \textbf{0.055} & \underline{0.238}    & \underline{0.110}    & 0.406          & 0.305          & 0.599          & 0.591          \\
FAN     & 0.189          & 0.066          & 0.257          & 0.128          & \underline{0.333}    & \underline{0.191}    & 0.513          & 0.474          \\ \hline
AEFIN   & \underline{0.180}    & \underline{0.061}    & \textbf{0.237} & \textbf{0.099} & \textbf{0.324} & \textbf{0.186} & \textbf{0.467} & \textbf{0.356} \\ \hline
\end{tabular}
} 
\vspace{0.15cm} 
\end{table}

\begin{figure}[H]
    \centering
    \includegraphics[width=1\linewidth]{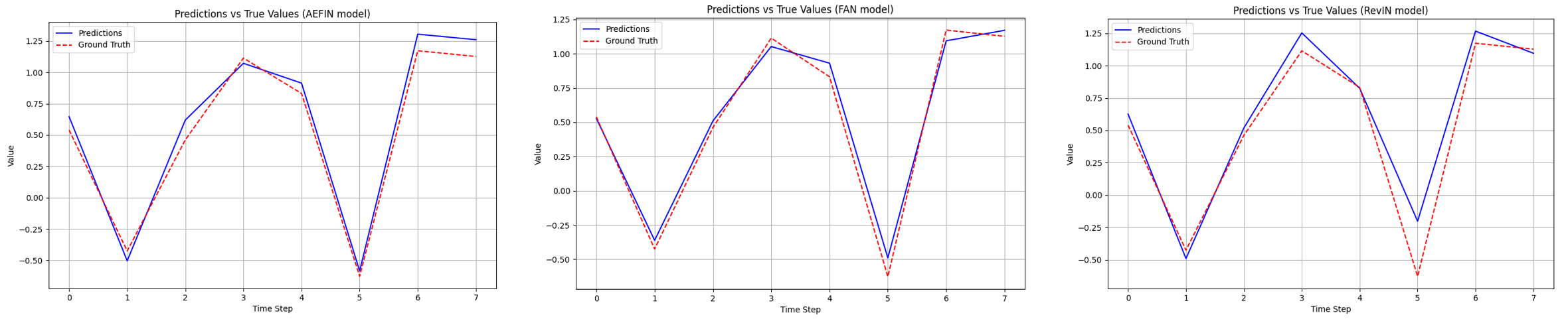}
    \caption{True and predicted values when the forecasting length is 96.}
    \label{fig:example3}
\end{figure}

\begin{figure}[H]
    \centering
    \includegraphics[width=1\linewidth]{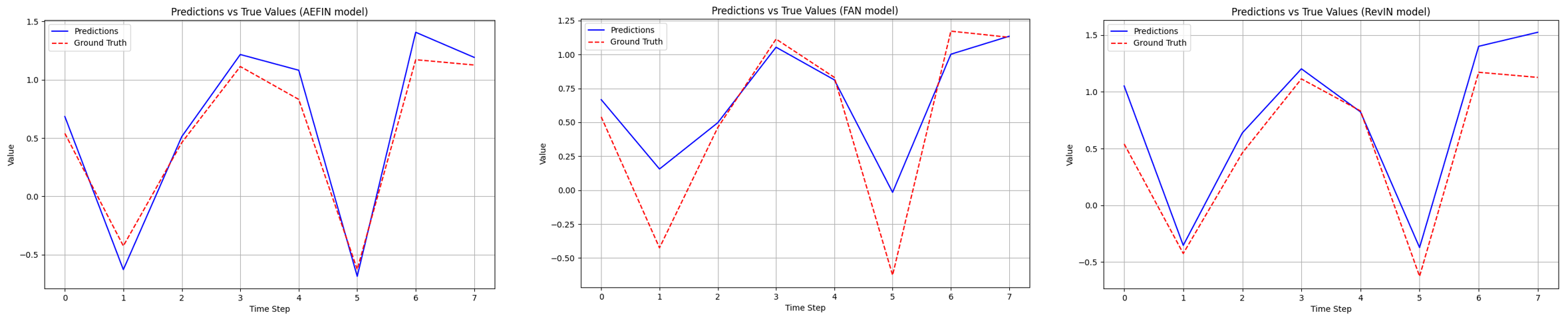}
    \caption{True and predicted values when the forecasting length is 168.}
    \label{fig:example4}
\end{figure}

\begin{figure}[H]
    \centering
    \includegraphics[width=1\linewidth]{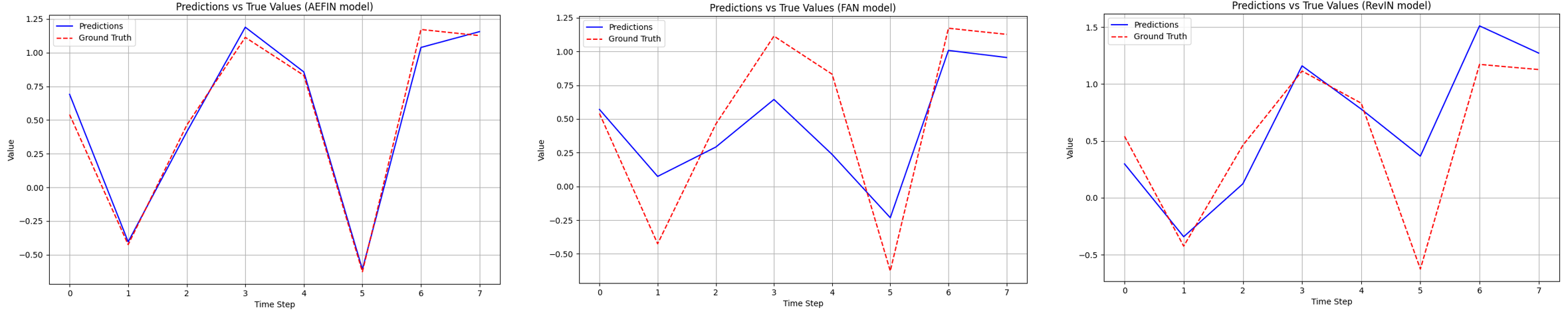}
    \caption{True and predicted values when the forecasting length is 720.}
    \label{fig:example5}
\end{figure}

Fig.\ref{fig:example3} to Fig.\ref{fig:example5} show the actual value and predicted value curves of our AEFIN model, FAN model, and RevIN model in the short, medium, and long term forecasting on the ExchangeRate dataset. It can be seen that these two models perform better in short-term forecasting. As the forecasting time increases, our model shows better forecasting results.

\section{Limitation}

\begin{figure}[H]
    \centering
    \includegraphics[width=1\linewidth]{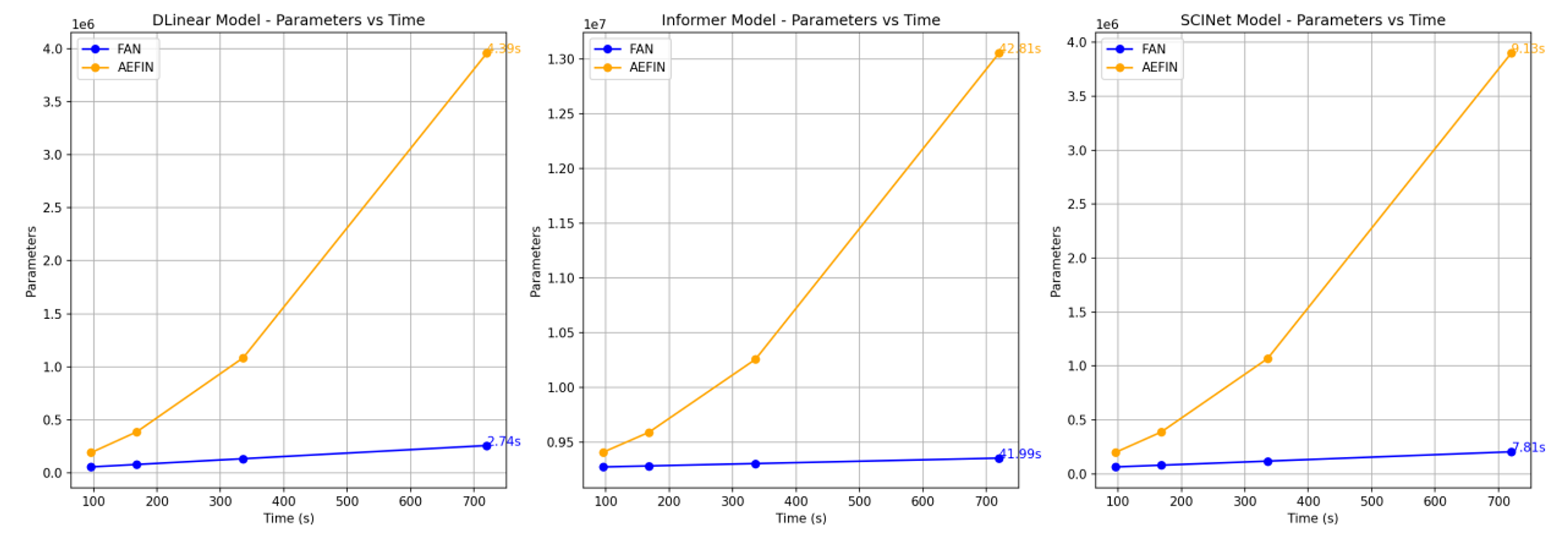}
    \caption{Parametric quantity curves for different models at various time steps.}
    \label{fig:example6}
    \label{fig:enter-label}
\end{figure}

Table~\ref{tab:example8} shows the number of parameters of our AEFIN model and FAN model at four different time steps and the average time spent on each round of training. Fig.\ref{fig:example6} shows the parameter change curves of the two models at each time step, from left to right: DLinear, Informer, and SCINet backbone. The yellow line indicates the number of parameters in our model and the blue line indicates the number of parameters in the baseline model.

\begin{table}[H]
\caption{The number of parameters versus average time spent per training round for different models at each time step.}  
\label{tab:example8} 
\centering  
\resizebox{8.9cm}{!}{  
\begin{tabular}{ccccccccc}
\hline
\multicolumn{3}{c}{Method}                            & \multicolumn{2}{c}{DLinear}       & \multicolumn{2}{c}{Informer}       & \multicolumn{2}{c}{SCINet}        \\
\hline
\multicolumn{3}{c}{metrics}                           & Parameter & Time                  & Parameter & Time                   & Parameter & Time                  \\ \hline
\multicolumn{2}{c}{\multirow{4}{*}{FAN}}   & 96       & 57840     & \multirow{4}{*}{2.74} & 9274168   & \multirow{4}{*}{41.99} & 63216     & \multirow{4}{*}{7.81} \\
\multicolumn{2}{c}{}                       & 168      & 81096     &                       & 9283456   &                        & 79416     &                       \\
\multicolumn{2}{c}{}                       & 336      & 135360    &                       & 9305128   &                        & 117216    &                       \\
\multicolumn{2}{c}{}                       & 720      & 259392    &                       & 9354664   &                        & 203616    &                       \\ \hline
\multicolumn{2}{c}{\multirow{4}{*}{AEFIN}} & 96       & 192473    & \multirow{4}{*}{4.39} & 9408801   & \multirow{4}{*}{42.81} & 197849    & \multirow{4}{*}{9.13} \\
\multicolumn{2}{c}{}                       & 168      & 388223    &                       & 9590583   &                        & 386543    &                       \\
\multicolumn{2}{c}{}                       & 336      & 1086893   &                       & 10256661  &                        & 1068749   &                       \\
\multicolumn{2}{c}{}                       & 720      & 3955661   &                       & 13050933  &                        & 3899885   &                       \\ \hline
\end{tabular}
} 
\end{table}

As can be seen from the figure, the parameters of our improved model are significantly higher than the baseline, and increase rapidly with the increase of forecasting time, which leads to greater consumption of computing resources, slower reasoning speed, and increased model complexity and maintenance difficulty.

Therefore, although the cross-attention mechanism improves the feature interaction ability of the model to a certain extent, how to perform efficient training and reasoning for ultra-large-scale datasets and complex time series problems is still an important challenge. In the future, it is possible to consider combining technologies such as distributed computing and graph neural networks to further improve the scalability and computing efficiency of the model.

\section{Conclusion}

This paper proposes a new time series forecasting framework, which aims to improve the forecasting accuracy of non-stationary time series data and fill the gap of existing methods in dealing with non-stationarity. In view of the limitations of traditional models in capturing the characteristics of non-stationary data, we have achieved significant improvements through the following three key innovations:

\subsubsection*{Introduction of Cross Attention Mechanism}
We apply the cross-attention mechanism in non-stationary time series forecasting to enhance the information sharing between the stable and non-stationary components in the sequence. Through this mechanism, the model can effectively transfer information between multiple time series components. This method improves the model's adaptability to non-stationary sequences.

\subsubsection*{Combination of Fourier Analysis Network and Multi-Layer Perceptron}
In order to enhance the model's ability to capture seasonal and trend characteristics in time series, we combined the Fourier analysis network with MLP to make full use of the Fourier analysis network. Information extraction of seasonal changes and trend changes by MLP. Through this cross-domain fusion, the model can more accurately model the seasonal and trend characteristics in time series data.

\subsubsection*{Design of Loss Function}
We propose a loss function that combines stability constraints in both time and frequency domains, which can guide the model to better adapt to the changing patterns of non-stationary data during training. Through this innovative loss function, the model can achieve a balance in the time domain and frequency domain during training, thereby improving forecasting accuracy, especially when the data has strong non-stationarity.

Experimental results verify the superiority of the proposed method on multiple standard datasets. Compared with existing mainstream methods, our framework has achieved significant improvements in forecasting accuracy, especially in practical applications in the financial field, and can provide more accurate and reliable forecasting results, providing more scientific basis for relevant decision-making support.

\vspace{12pt}

\end{document}